\documentclass{article}

\usepackage{arxiv}

\usepackage[utf8]{inputenc} 
\usepackage[T1]{fontenc}    
\usepackage{hyperref}       
\usepackage{url}            
\usepackage{booktabs}       
\usepackage{amsfonts}       
\usepackage{nicefrac}       
\usepackage{microtype}      
\usepackage{lipsum}
\usepackage{array}
\usepackage{graphicx}

\title{Theory-Guided Machine Learning for Process Simulation of Advanced Composites}

\author{
  Navid Zobeiry \\
  Department of Materials Sciences and Engineering\\
  University of Washington\\
  Seattle, United States \\
  \texttt{navidz@uw.edu} \\
   \And
  Anoush Poursartip \\
  Composites Research Network\\
  The University of British Columbia\\
  Vancouver, Canada \\
  \texttt{anoush.poursartip@ubc.ca} \\
}

\begin{document}
\maketitle

\begin{abstract}
Science-based simulation tools such as Finite Element (FE) models are routinely used in scientific and engineering applications. While their success is strongly dependent on our understanding of underlying governing physical laws, they suffer inherent limitations including trade-off between fidelity/accuracy and speed. The recent rise of Machine Learning (ML) proposes a theory-agnostic paradigm. In complex multi-physics problems, however, creating large enough datasets for successful training of ML models has proven to be challenging. One promising strategy to bridge the divide between these approaches and take advantage of their respective strengths is Theory-Guided Machine Learning (TGML) which aims to integrate physical laws into ML algorithms. In this paper, three case studies on thermal management during processing of advanced composites are presented and studied using FE, ML and TGML. A structured approach to incrementally adding increasingly complex physics to training of TGML model is presented.  The benefits of TGML over ML models are seen in more accurate predictions, particularly outside the training region, and ability to train with small datasets. One benefit of TGML over FE is significant speed improvement to potentially develop real-time feedback systems. A recent successful implementation of a TGML model to assess producibility of aerospace composite parts is presented. 
\end{abstract}


\section{Introduction}
In the context of scientific and engineering problems, it becomes critical to discover underlying physical laws and predict the effects of controllable parameters on desirable physical outcomes. We can argue that the major focus of scientific and engineering research over the last century has been to develop a body of theory that can explain and predict physical phenomena. It has been proposed that one can define the traditional stages (or paradigms) of scientific development as moving from empirical science to model-based theoretical science through to computational science[1]. With the recent rise of Machine Learning (ML), the fourth paradigm based on data-driven modeling is being explored in various branches of science and engineering[1].

Generally speaking, multi-physics problems can be expressed mathematically in terms of a series of Partial Differential Equations (PDEs). Typically these can be obtained on the basis of conservation laws such as mass and momentum conservations, energy conservation or equilibrium[2]. In specific cases, these PDEs may be solved analytically to discover underlying physical correlations and establish closed-form solutions (e.g. 1D steady-state heat transfer problem[3]). However, for general cases, high-fidelity analytical solutions cannot be obtained. Instead, these PDEs are solved using established numerical methods such as the Finite Element (FE) method with a rigorous mathematical foundation[4], or in more recently using Physics-informed Machine learning (PIML) method[5–7]. Processing of fiber reinforced polymer composites (i.e. advanced composites), for example, is a complex and multi-physics problem of heat and mass transfer, thermo-chemical phase transitions, and highly nonlinear and time-dependent viscoelastic stress developments[8–14]. For advanced composites, the application of theory through modern FE methods in industries such as aerospace and automotive, has enabled a wide range of advances in material development, manufacturing and design[15–21].

\begin{figure}
  \centering
  \includegraphics{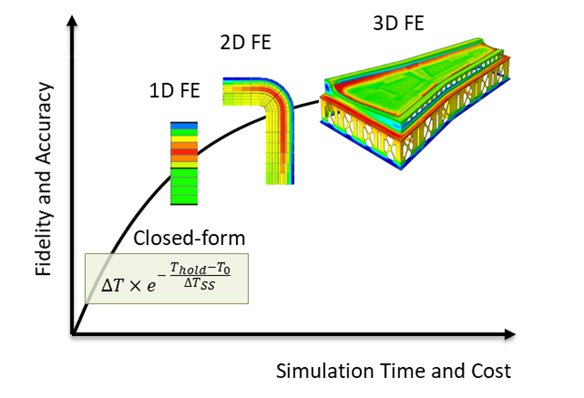}
  \caption{Trade-off between fidelity/accuracy and simulation time for different solutions for thermo-chemical analysis of composite parts during processing (FE simulation figures are provided by Convergent Manufacturing Technologies and used by permission in this article).}
  \label{fig:fig1}
\end{figure}

However, while acknowledging the tremendous success of science-based simulation methods, there are limitations with these approaches. One notable limitation is the trade-off between fidelity/accuracy and development/simulation time. Developing high fidelity FE tools requires tremendous efforts on characterization, verification and validation. Once fully developed, high accuracy comes at the price of characterization time, set-up time and simulation time. Take thermo-chemical analysis of advanced composites during processing as an example. For a typical aerospace composite part (e.g. the wing skin of the Boeing 787), full 3D thermo-chemical analysis can be conducted using validated commercial FE tools in the span of several hours or days[10, 16, 22, 23]. Considering the dominant heat transfer mechanism in a thin part, much faster but lower fidelity 1D/2D FE tools or even closed-form solutions may be used[8, 24]. In some applications such as conceptual design phase, these tools may offer acceptable accuracies at much higher speeds (e.g. real-time for closed-form, seconds for 1D and minutes for 2D). This is schematically shown in Figure \ref{fig:fig1} for processing of a representative aerospace composite part where usually gain in accuracy is incremental while simulation time increases at a much higher rate. For such a complex and multi-physics problem, closed-form solutions do not provide acceptable accuracies[24]. However, 1D/2D simulations have shown to be quite effective for exploring the design spectrum in the conceptual design phase[17, 19, 25, 26], followed by 3D simulations for detailed studies[27, 28].

As opposed to science-based simulation approaches that start by establishing governing PDEs, theory-agnostic data-driven modeling relies on physical data to discover underlying correlations and hidden governing laws[2]. This highlights the importance of data generation and data quality to successfully employ ML in specific engineering applications. For manufacturing problems, interestingly, the same advances in computer technology that have enabled computational science, have led to equally significant advances in sensing and automation.  This has led to major improvements in industrial capability, now often called Industry 4.0, where large amount of physical data can be generated via technologies such as the Internet of Things (IoT)[29].

From the perspective of those grounded in science, the onslaught of capabilities in ubiquitous sensing, big data, AI (including ML) and so forth presents an interesting dilemma. Whole new disciplines such as data science have grown very fast just in the last decade. We note that, despite the significant success of ML in areas such as image and pattern recognition, the implementation of black–box data science concepts have so far achieved limited penetration and success in scientific and engineering domains. Even though many successful examples have been demonstrated in physical problems such as high-fidelity turbulence modelling using Random Forest and Deep Neural Network (DNN) algorithms[30, 31], and analysis of complex microscopy data using k-Means Clustering and Principal Component Analysis (PCA)[32], this is still a far cry from the significant success seen with ML in non-engineering applications.

There are several reasons that have limited the application of ML in physical domains.  Even with recent advances in sensing and automation, data in engineering problems are inherently small and fragmented and often suffer from underrepresented classes[33]. Oftentimes, due to cost considerations, physical constraints or complexity of measurements, Key Performance Indicators (KPI) such as material strength or microstructure can only be measured from a limited number of tests. In addition, physical variables usually demonstrate highly nonlinear and time dependent responses, which when combined with the lack of big data, cause ML to fail in representing the complex physics of the problem. Other issues such as lack of interpretability of ML models and superficial/physically inconsistent correlations particularly near or outside the boundaries of training zones make it less desirable to implement black-box ML approaches.

Considering limitations in both science-based simulation approaches and data-driven modeling methods, the question of interest is how to combine these methods to take advantage of their respective strengths. In contrast to theory-agnostic approaches in data science, Theory-Guided Machine Learning (TGML) is a new and emerging paradigm that integrates existing domain knowledge (implicit or explicit including FE simulation results) in machine learning to introduce scientific consistency in solutions[2, 25, 33–35]. This includes a variety of approaches such as generating data with FE tools, introducing physics-based features, establishing physically interpretable model architectures, enriching loss functions with physics-based functions, and constraining response surfaces[33]. TGML can be used in engineering applications including the following:

\begin{itemize}
\item Developing surrogate ML models using numerical data from high fidelity theory-based FE simulation tools[36].  These surrogate ML models can be used to significantly speed up the set-up time and run-time of simulation and potentially develop real-time feedback systems for engineering applications.
\item Developing high fidelity predictive ML models based on available approximate solutions and limited number of physical or accurate numerical data points. In turbulence modeling, for example, TGML has been proposed to increase the accuracy of the Reynolds-Averaged Naiver-Stokes (RANS) models based on limited data from high fidelity Direct Numerical Simulation (DNS) tools[31]. 
\item Data preparation, analysis and visualization of large and complex datasets (e.g. complex 3D micro-CT images or C-scan Ultrasonic Images), and numerical data (e.g. complex results from Molecular Dynamics) to identify underlying correlations and distill physical insight. These correlations can be used to infer causations based on existing domain knowledge.  An example in the composites field is the development of neural net-based cure kinetics model for composites processing[37, 38].
\item Design optimization using fast and high-fidelity ML models. For example, such an approach can be used to optimize manufacturing cycles by establishing relationships between process parameters, process-induced defects and end-part performance. 
\end{itemize}

This paper will demonstrate how challenges in employing black-box data science approaches and limitations in science-based simulation methods can be addressed with the capabilities of a TGML approach. Several case studies focused on thermal management during manufacturing of aerospace composite parts are presented. The aerospace industry increasingly uses science-based FE simulation tools to design and optimize processing cycles and mitigate defects for advanced composites used to manufacture very large unitized structures in the latest generation aircraft[10, 16, 22, 23, 39–41].  Three case studies for thermo-chemical analysis of composites are considered to take advantage of TGML to speed up existing FE simulation tools and also to develop high fidelity models from approximate closed-form solutions. FE, black-box ML and TGML approaches are used in each case study. As the complexity of cases is increased, increasingly complex physics-based TGML techniques are introduced to overcome traditional ML challenges. Finally, a recent successful implementation of a TGML model in a commercial simulation tool is presented with significant gain in simulation speed compared to traditional FE tools.

\section{Background: Process Simulation of Advanced Composites}

In the aerospace industry, advanced composites are used due to their extremely high specific stiffness and strength ratios. In the Boeing 787 and Airbus A350, for example, thermoset-based resin systems reinforced with carbon fibers are used to fabricate primary structural elements including the wing and fuselage. These large and complex parts are cured in massive autoclaves where temperature and pressure cycles are tightly controlled. During this process, every point in the structure must see a temperature cycle within an upper and lower bound, as the final properties are highly dependent on this process history[42]. The effects of the temperature history on evolution of material properties and consequently manufacturing-induced defects are governed by complex and highly nonlinear laws[8, 11, 12]. In order to predict the temperature distribution in a composite part, heat transfer analysis is performed. However, since curing of thermoset-based composites is exothermic, heat transfer analysis should include the effect of heat generation in the resin system. The general 3D heat equation for such a case can be written as follows[8]:

\begin{equation}
\frac{\partial}{\partial t} (\rho C_P\ T)=\frac{\partial}{\partial x} (k_{xx}\frac{\partial T}{\partial x})+\frac{\partial}{\partial y} (k_{yy}\frac{\partial T}{\partial y})+\frac{\partial}{\partial z} (k_{zz}\frac{\partial T}{\partial z})+\dot{Q}
\end{equation}

In which \(T\) is the temperature, \(\rho\) is the part density, \(C_P\) is the part specific heat capacity, \(k\) is the part conductivity, and \(\dot{Q}\) is the rate of heat generation in part due to the exothermic polymerization reaction. Heat generation in a composite part can be calculated as:

\begin{equation}
\dot{Q}=\frac{\partial\chi}{\partial t}(1-V_f\ )\ \rho_R\ H_R
\end{equation}

Where \(\chi\) is the resin degree of cure, \(V_f\) is the fiber volume fraction, \(\rho_R\) is the resin density and \(H_R\) is the resin total heat of reaction. For typical aerospace thermoset-based resin systems, the cure rate is governed by both kinetic and diffusion mechanisms:

\begin{equation}
\frac{\partial\chi}{\partial t}=(\frac{1}{\frac{\partial\chi_{Kinetic}}{\partial t}}+\frac{1}{\frac{\partial\chi_{Diffusion}}{\partial t}})^{-1}
\end{equation}

The kinetic effect itself can also be very complex, but is typically represented using Arrhenius-based equations[39], e.g.

\begin{equation}
\frac{\partial\chi_{Kinetic}}{\partial t}=Ae^{\frac{-\Delta E}{RT}}\chi^m (1-\chi)^n
\end{equation}

Where \(\Delta E\) is the activation energy, \(R\) is the gas constant, and \(A\), \(m\) and \(n\) are constants. Similarly, the diffusion term is usually defined as[43]:

\begin{equation}
\frac{\partial\chi_{Diffusion}}{\partial t}=k_d\ e^{\frac{-B}{a_f\ (T-T_g\ )+b_f}}
\end{equation}

Where \(T_g\) is the glass transition temperature, and \(k_d\), \(B\), \(a_f\) and \(b_f\) are material constants. The key point is that the rate of heat evolution during cure is a function of material properties and local temperature history and is therefore highly non-linear in space and time.  In convective heating of composite parts, during heat-up and while the cure rate is still low, part temperature lags behind the air temperature[8, 42]. As the polymerization reaction starts, the heat of reaction may cause the part temperature to increase beyond that of the air temperature. To control the quality of the part, both minimum part temperature (i.e. thermal lag) and maximum part temperature (i.e. exotherm) have to be controlled tightly.

It has been shown that the complex 3D system of a curing part on a tool can be reasonably simplified by a 1D representation of the through-thickness thermal response in critical zones[19, 24].  Therefore, to predict the part thermal lag and exotherm, Equation 1 is simplified such that the temperature gradient is only in the z direction:

\begin{equation}
\frac{\partial}{\partial t} (\rho C_P\ T)=\frac{\partial}{\partial z} (k\frac{\partial T}{\partial z})+\dot{Q}
\end{equation}

\begin{figure}
  \centering
  \includegraphics{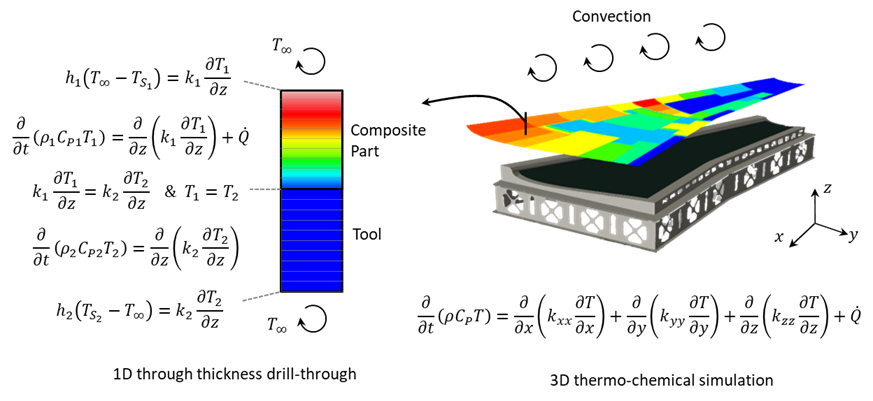}
  \caption{Governing equations for thermo-chemical analysis of exothermic composite parts in 3D and equivalent 1D drill-through (3D FE simulation figure is provided by Convergent Manufacturing Technologies and used by permission in this article).}
  \label{fig:fig2}
\end{figure}

In which \(\dot{Q}\)  is defined in Equations 2 to 5. Systems of equations (3D and equivalent 1D), for convective heating and curing of a composite part on a tool are shown in Figure \ref{fig:fig2}. Here, the system of equations includes convective heat transfer from air to part and tool, conductive heat transfer in part and tool, and heat generation in part. In the 1D equations in Figure \ref{fig:fig2}, subscript 1 refers to part thermal properties and subscript 2 refers to tool thermal properties. These can be solved by a variety of methods, including general-purpose commercial finite element simulation tools such as ABAQUS or ANSYS with the addition of user materials, or specialized ones such as COMPRO[44] or PAM-COMPOSITES[45] with built-in material libraries. Tailored packages, such as RAVEN[46], fully automate the 1D representation and solution of the problem.  It should be noted that currently RAVEN and COMPRO packages are widely used in aerospace industry. For example, these simulation packages were previously used to design processing cycles of composites parts including fuselage sections of Boeing 787. Validation of these tools are extensively discussed in the literature[28, 43, 47]. In this study, RAVEN software is used for all FE simulations. These simulations require full characterization of thermo-chemical properties of materials which is briefly discussed next.

\subsection{Material Characterization}
In this study, an aerospace grade composite material, AS4/8552 by HEXCEL is considered in all case studies[48]. 8552 is a toughened epoxy system which is a blend of epoxy and PES (20-30\% wt.)[9], and AS4 is a high strength carbon fiber with a dimeter of about 7 µm[49]. The composite part has a fiber volume fraction of about 57\%[9, 43]. The thermo-chemical properties of this material have been extensively studied in the literature and are considered standard material properties[43]. This includes cure kinetics, specific heat capacity, thermal conductivity and density[9, 40, 43, 50]. Cure kinetics characterization, for example, was previously conducted by isothermal and dynamic DSC (Dynamic Scanning Calorimetry) tests[43]. Tooling materials used in this study are Invar (Case 2 and Case 3) and nominal tooling composite (Case 1). Thermo-chemical properties of typical Invar and composite tooling materials used in aerospace industry, including specific heat capacity, density and thermal conductivity, are also studied and reported in the literature[24, 26]. These material properties (both curing composites and tooling) are implemented in the commercial software RAVEN as standard material models which are used in this study.

\section{Theory-Guided Machine Learning for Processing of Composites}
In this section, we first demonstrate the limitations of theory-agnostic ML approaches for composite processing. We then use TGML to introduce physical consistency in ML models and show it is possible to reduce maximum prediction errors to a reasonable level. Finally, TGML is used to develop surrogate ML models to significantly speed up FE simulations such that they can be used for real-time feedback systems.  Three case studies are considered, with the governing equations becoming increasingly more complex, see Figure 3:

\begin{itemize}
\item Case I: Assessment of quasi steady-state thermal lag in an inert tool.  The starting TGML strategy is the selection of physics-based feature transformations based on a closed-form solution.
\item Case II: Prediction of the maximum transient thermal lag in a curing composite part on a metallic tool.  The TGML strategy uses a physics-based rationale for choice of activation function, and pre-training using a physics-based approximate solution.
\item Case III: Prediction of the maximum transient exotherm in a curing composite part on a metallic tool.  The TGML strategy is extended to this highly non-linear problem using understanding of the asymptotic behavior of the physics-based solution.
\end{itemize}

\begin{figure}
  \centering
  \includegraphics{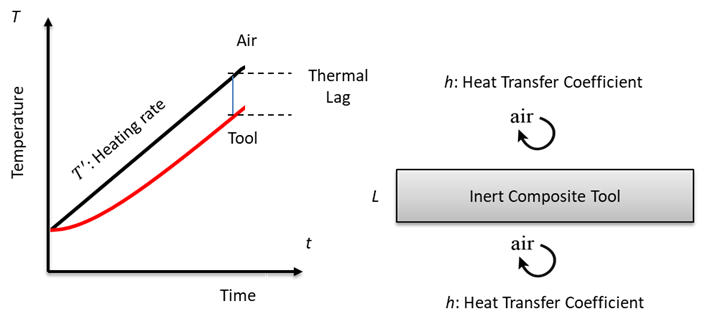}
  \caption{A 1D representation of an inert composite tool subjected to convective heating with a constant heating rate and quasi steady-state thermal lag in the tool.}
  \label{fig:fig3}
\end{figure}

\subsection{Analysis Details}
For all cases, 1D FE simulations were conducted using RAVEN Version 3.9.2 software[46] to generate the datasets.  For all simulations, input variables were selected randomly from predefined physically consistent ranges, and prediction values were recorded accordingly. After creating datasets, the data was split into two parts, 70\% training and 30\% validation. Testing datasets for examining underlying trends were created separately. Feed-forward dense Neural Network (NN) models (i.e. no cyclic connections between nodes) with Fully Connected (FC) layers were trained using the high level API in TensorFlow (version 1.8.0)[51], Estimator, available in Python (version 3.6.8). Before each training, features and outputs were normalized. After each training, three values were recorded: maximum prediction error, Root Mean Square Error (RMSE) and Normalized Root Mean Square Error (NRMSE). Over-fitted models were rejected based on the NRMSE difference between training and validation sets with a threshold of 0.02. This value was selected since it corresponds to an error of about ±1.0 ℃.  For heat transfer applications, this is close to ±1.1 °C which is the Special Limits of Error (SLE) for a type J thermocouple used in the aerospace industry. RMSE and NRMSE are defined below:

\begin{equation}
RMSE=\sqrt{\frac{\sum^{N}{(T_i-P_i)^2}}{N}}\ \ \ and \ \ NRMSE=\frac{RMSE}{T_{max}-T_{min}}
\end{equation}

In which \(T_i\) is the output temperature, \(P_i\) is the predicted temperature by the trained model and N is the total number of data points.  \(T_{min}\) and \(T_{max}\) are minimum and maximum temperatures in the dataset respectively. Maximum prediction error was defined for the validation set as below:

\begin{equation}
Maximum Error=max(|T_i-P_i|)
\end{equation}

Selection of hyper-parameters in each case study was based on a grid search to reduce RMSE in the validation set. The following parameters were varied for the grid search:
\begin{itemize}
\item \textit{Neural Network architecture}: Several architectures were tried with 3 to 5 hidden layers and 5 to 50 nodes per hidden layer. Results from three specific architectures are compared in Case 1: small NN (3 hidden layers with 5 nodes per layer), medium NN (4 hidden layers with 10 nodes per layer), and large NN (5 hidden layers with 25 nodes per layer).
\item \textit{Activation function}: Rectified Linear Unit (ReLU)[52], sigmoid and tanh[51]. This is further discussed in Case 2 where results from ReLU and tanh functions are compared.
\item \textit{Loss function}: Mean Squared Error (MSE) and Mean Absolute Error (MAE). After grid search, MSE was used in trainings. As mentioned before, RMSE is reported for trainings.
\item \textit{Optimization}: Gradient Descent (GD) without regularization, Adaptive Gradient Descent (AGD) without regularization, Proximal Gradient Descent (PGD) with regularization, Proximal Adaptive Gradient (PAG)[53] with regularization and Adam[54] without regularization. After grid search, GD and PAG were selected. Overall PAG showed to be the best optimizer in most cases discussed in this paper. Learning rates varied between 0.0005-0.01 in GD and 0.02-0.06 in PAG. Errors are compared in Case 3.
\item \textit{Regularization}: Least Absolute Shrinkage Selector Operator (LASSO) and Ridge[55, 56] with learning rates of 0.0, 0.001, 0.01, 0.1 were for PGD and PAG optimizers. Results are discussed in Case 1. 
\item \textit{Batch size}: 20, 40, 50, 60 and 100. A batch size of 50 was selected in all cases after grid search.
\item \textit{Normalization/Balancing}: Three different functions were tried:
\begin{equation}
\bar{x}=\frac{x-\mu}{\sigma},\ \ \bar{x}=\frac{x-x_{min}}{x_{max}-x_{min}},\ \ \ \bar{x}=2\frac{x-x_{min}}{x_{max}-x_{min}}-1
\end{equation}

Here, \(\mu\) is the mean value, \(\sigma\) is the standard deviation, \(x_{min}\) is the minimum value and \(x_{max}\) is the maximum value in a dataset. For all cases, it was determined that the last function with normalized features between -1 and 1, gives the highest accuracy with the lowest RMSE in validation set. 
\item \textit{Training Steps or Iterations}: Between 20,000 - 250,000 until validation error became negligible. More details are discussed in each case. It should be noted that iteration is different than epoch. Each epoch has several iterations equal to size of dataset divided by batch size. Therefore, total number of iterations is equal to total number of epochs multiplied by iterations per epoch.
\item \textit{Data size}: Between 100 - 10,000 as discussed in each case.
\item \textit{Initialization}: Random normal, He normal, Glorot uniform (Xavier)[57] and physics-based initialization as discussed in Case 2. For all cases, Xavier initialization was selected after the grid search. For Case 2, it is shown that physics-based initialization (i.e. pre-training) significantly improves the accuracy of the trained models. 
\end{itemize}

\subsection{Case Study 1: Quasi Steady-State Lag in an Inert Composite Tool}
Consider an inert composite tool (through-thickness conductivity of \(0.5 \ W/mK\) and diffusivity of \(0.5\times{10^{-6}}  \ m^2/s)\)[24] with a thickness of L which is heated via convection. A 1D representation of the problem is schematically depicted in Figure 3. Assume that the heating rate, \(T^\prime\), is constant[48]. Also assume that heat transfer coefficient, h, is uniform and constant above and below the tool[24]. As the air temperature is increased at a constant rate, the center of the tool heats up at a slower rate initially. Eventually a quasi-steady-state is reached where the difference between the air temperature and tool temperature remains constant[8]. It is of interest to predict the value of this steady-state thermal lag, \(\Delta T\). 10,000 1D FE simulations were conducted to create a dataset. In each simulation, the tool thickness, \(L\), was varied between 2 and 20 mm, the heat transfer coefficient, \(h\), between 20 and 100 \(W/(m^2 K)\)[8],  and the heating rate, \(T^\prime\), between 1 and 5 °C/min[48]. The inputs were generated randomly for each simulation. NumPy.random.rand in Python was used to generate random inputs with uniform distributions within specified ranges. 

A NN model with 3 hidden FC layers and 5 nodes per layer was initially chosen for training as depicted in Figure 4a. The ReLU activation function and Gradient Descent (GD) optimization scheme with a slow learning rate of 0.001without any regularization were used to train the model. After 50,000 iterations RMSE became almost constant. Many trainings were performed where random numbers of data points between 100 and 10,000 were selected. Errors from validation sets are shown in Figure 5 as a function of the dataset size. A trend can be observed where error decreases as the size of the dataset increases. Eventually the NRMSE becomes stable as the dataset becomes larger than about 1,500 data points.

\begin{figure}
  \centering
  \includegraphics{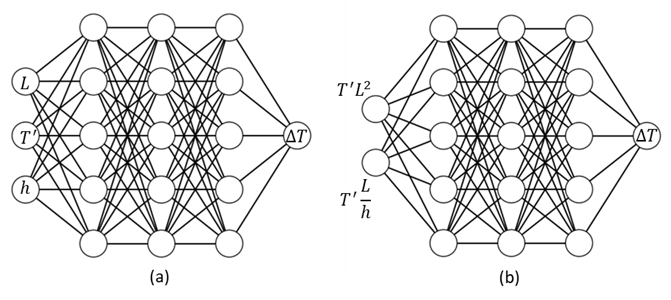}
  \caption{(a) A neural network model with fully connected (FC) layers used in case study 1 to predict thermal lag using 3 original features and, (b) A neural network model to predict thermal lag using two physics-based features.}
  \label{fig:fig4}
\end{figure}

\begin{figure}
  \centering
  \includegraphics{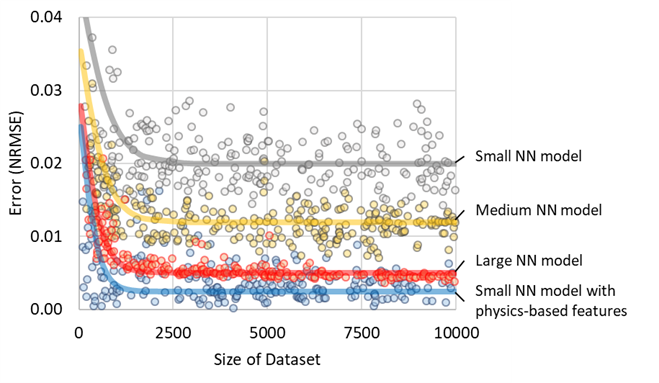}
  \caption{Normalized error (NRMSE) of different NN models as a function of dataset size in case study 1. Number of hidden layers, nodes and error values are listed under Table 1.}
  \label{fig:fig5}
\end{figure}

The same study was repeated with two different architectures: a medium size NN model with 4 hidden layers and 10 nodes per layer and a large NN model with 5 hidden layers and 25 nodes per layer. For the medium size model, 100,000 iterations and for the large model 200,000 iterations were performed. Results are compared in Figure 5. Similar trends as the small NN model are observed. However, as the model complexity is increased, the magnitude of error drops as listed in Table 1. It should be noted that even though NRMSE and RMSE drop and become small (0.004 and 0.3 °C respectively), the maximum error value does not reduce below 3.2 °C in the largest trained NN model.  Note that maximum error is a more important metric than RMSE as it goes to confidence in using an ML model.  Given that cure cycle specifications are typically based on ±5 °C limits and given that type J thermocouples have errors in the range of ±1.1 °C, a maximum prediction error of 3.2 °C is higher than desirable.

\begin{table}
\caption{Mean and max errors to predict thermal lag in case study 1 after training NN models with 10,000 data points and Gradient Decent (GD) optimizer with a learning rate of 0.001}
\centering
\begin{tabular}{p{0.20\linewidth} | >{\centering\arraybackslash}p{0.1\linewidth} | >{\centering\arraybackslash}p{0.1\linewidth} | >{\centering\arraybackslash}p{0.1\linewidth} | >{\centering\arraybackslash}p{0.1\linewidth} | >{\centering\arraybackslash}p{0.1\linewidth} | >{\centering\arraybackslash}p{0.1\linewidth}}
\hline
\textbf{Model} & \textbf{Optimizer} & \textbf{Layers} & \textbf{Nodes/ Layer} & \textbf{NRMSE} & \textbf{RMSE (°C)} & \textbf{Max Error (°C)} \\ \hline
Small NN & GD 50K iterations & 3 & 5 & 0.020 & 1.5 & 12.2\\
Medium NN & GD 100K iterations & 4 & 10 & 0.011 & 0.7 & 4.7\\
Large NN & GD 200K iterations & 5 & 25 & 0.004 & 0.3 & 3.2\\
Small NN with physics-based features & GD 50K iterations & 3 & 5 & 0.002 & 0.1 & 1.8\\\hline
\end{tabular}
\label{tab:table1}
\end{table}

The steady-state heat transfer problem has been studied extensively in the literature[3]. For such a problem, it is known that non-dimensional heat transfer numbers can effectively describe the thermal lag in the system. Specifically for the quasi-steady-state lag in convective heating, it has been shown that the following two non-dimensional numbers govern the problem[24, 58]:

\begin{equation}
\frac{1}{Bi\times Fo}=\frac{1}{\frac{hL}{k}\times \frac{at}{L^2} } \propto \frac{T^\prime L}{h} \  \ \ and \  \ \ \frac{1}{Fo}=\frac{1}{\frac{at}{L^2} } \propto T^\prime L^2
\end{equation}

In Equation 10, \(Fo\) is the Fourier number, \(Bi\) is the Biot number, a is the thermal diffusivity, \(k\) is thermal conductivity, \(t\) is time, and \(T^\prime\) is temperature rate[3, 58].  We now repeat the trainings using the above two physics-based features instead of the original features. We also use the small NN model architecture as shown in Figure 4b. The comparison of results is shown in Figure 5. It can be observed that small models which were trained using physics-based features outperform all the other models including large NN models. The RMSE reduces to 0.1 °C and the maximum error drops to 1.8 °C (Table 1), which is an acceptable value. By comparison, the special limits of error (SLE) of type J thermocouples typically used in industry is ±1.1 °C.
Aside from reducing maximum error, another important factor in developing confidence in the ML model as an engineering design tool is the underlying trend. To compare the capability of trained models to identify trends, a simple test case was considered where the heat transfer coefficient and thickness were kept constant (50 \(W/(m^2 K)\) and 10 mm  respectively) while the heating rate was varied between 0.25 and 10 °C/min. It should be noted that the original training was performed in the range of 1 to 5 °C/min.  Additional FE simulations were performed to compare with predictions of trained NN models as shown in Figure 6.  Non-TGML NN models, independent of the size of the dataset, perform poorly outside the training zone, in contrast to the TGML model, which captured the underlying trends well beyond the bounds of the training zone. 

\begin{figure}
  \centering
  \includegraphics{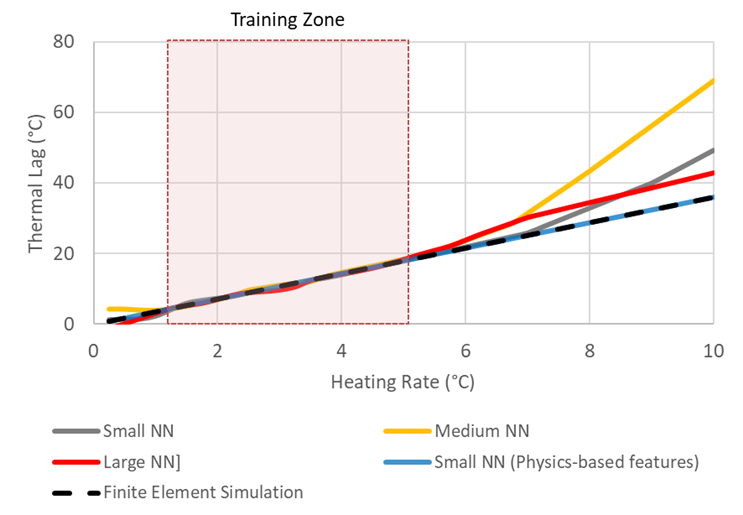}
  \caption{Predictions of the underlying trend beyond the training zone using different NN models compared with finite element simulation results in case study 1.}
  \label{fig:fig6}
\end{figure}

This is of course the consequence of integrating domain-knowledge in ML by introducing physics-based features. Although one might try to employ dimensionality reduction algorithms such as Principal Component Analysis (PCA)[59] to extract these physics-based features with trial and error, it would be a time-consuming process which may not result in exact features.  In any case, this trial and error approach becomes more cumbersome as the number of original features increases. One could argue that without knowing the theory, and the exact combination of non-dimensional numbers, arriving at the feature transformations discussed above would be quite unlikely. It should also be noted that feature transformation based on non-dimensional numbers is only applicable to this simple steady-state example. In the other transient case studies discussed below, more complex feature transformations are used. In fact, the blind application of non-dimensional numbers to transient problems might results in high errors[60].

It should be noted that the noise level in validation errors is mostly observed in the smallest model. As the model becomes larger, or as physics-based features are added to the model, noise level drops significantly. This is because smallest model does not have enough nodes and hidden layers to capture the full complexity of the physical problem. As random datasets are created and used for trainings, validation error varies depending on how much of the physics is captured during training. To show that this is due to a small NN architecture and not a regularization problem, results from grid search on regularization is presented here. Several trainings were performed with different regularization schemes as shown in Figure 7. This is based on validation RMSE using Gradient Descent optimization (learning rate = 0.001) and different combinations of regularization: no regularization, LASSO regularization (learning rate = 0.01), and Ridge regularization (learning rate = 0.01). This figure shows that using regularization, noise in validation RMSE in the smallest NN does not change but the average error increases slightly. This is because these regularization schemes may remove nodes from training to further reduce the capability of the model to capture the complexity of the problem. The largest NN model with no regularization, shows much lower noise level. 

\begin{figure}
  \centering
  \includegraphics{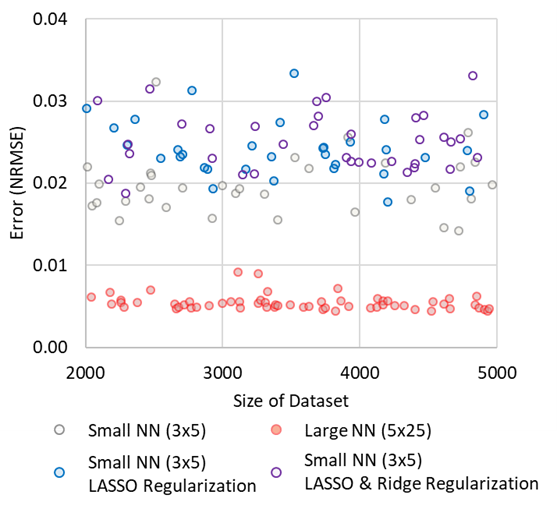}
  \caption{Effect of regularization on validation errors for Case 1: Gradient Descent optimization (learning rate=0.001), with or without LASSO regularization (learning rate=0.01) and Ridge regularization (learning rate=0.01).}
  \label{fig:fig7}
\end{figure}

\begin{figure}
  \centering
  \includegraphics{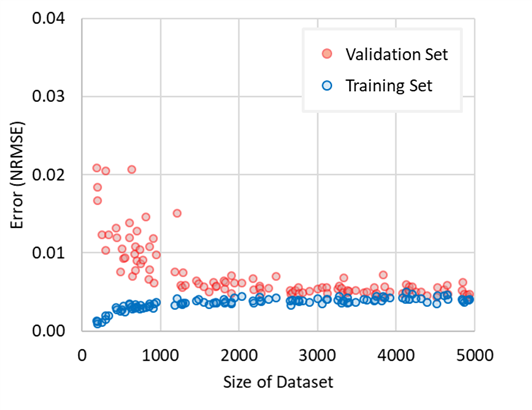}
  \caption{Comparison of training and validation errors as a function of dataset size for the large NN (5×25 nodes) in Case 1.}
  \label{fig:fig8}
\end{figure}

Figure 8 compares the RMSE in validation and training sets for the largest NN model. For small datasets, this shows some overfitting but the difference in RMSE is still acceptable (equivalent to less than 1℃ in predictions). As the dataset becomes larger, the difference between validation and training RMSEs becomes negligible. 

\subsection{Case Study 2: Transient Thermal Lag in a Curing Composite Part}
Consider a more complex problem where an uncured composite part with a thickness of \(L_1\), is placed on a metallic Invar tool with a thickness of \(L_2\). The part is cured by convective heating using a one-hold temperature cycle as schematically depicted in Figure 9.  It is generally true that airflow above and below the tool is different, and therefore we assume that the heat transfer coefficient above the part, h1, is different from the heat transfer coefficient under the tool, h2. A 1D representation of the problem is shown in Figure 9.  For this case study, we consider HEXCEL AS4/8552 carbon fiber/epoxy resin[48] with a known cure kinetics response[43]. As shown in Figure 9, the temperature cycle consists of a heat-up step from room temperature to 180°C with a constant heating rate of \(T^\prime\).  Air temperature is then kept constant for two hours before cooling down. During the heat-up step, initially the part temperature lags behind the air temperature. However, as the exothermic reaction initiates, the generated heat increases the part temperature beyond the air temperature. Here we aim to find the thermal lag, which is defined as the maximum difference between air temperature and any point in the part thickness during the heat-up step. This thermal lag is schematically shown in Figure 9. In contrast to the previous case study, this thermal lag is a function of the exothermic reaction and is a transient response. To make the problem more challenging, the FE solver was used to create a dataset with only 300 data points. Physically-consistent limits were imposed for part thickness, \(L_1\), between 2 and 20 mm; tool thickness, \(L_2\), between 8 and 20 mm; heat transfer coefficients, \(h_1\) and \(h_2\), between 20 and 100 \(\frac{W}{m^2K}\); and heating rate, \(T^\prime\), between 1 and 5 °C/min.

\begin{figure}
  \centering
  \includegraphics{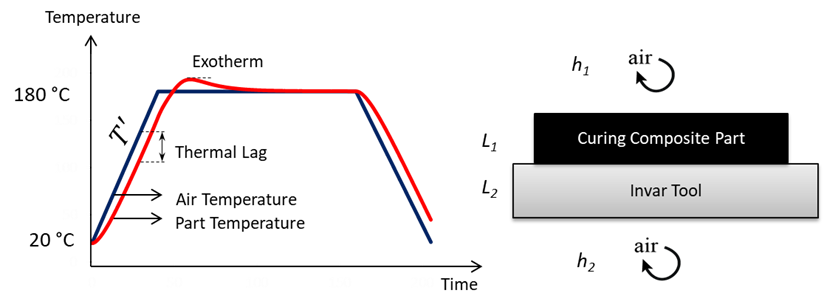}
  \caption{A 1D schematic representation of a curing composite part placed on a metallic Invar tool and subjected to a simple one-hold temperature cycle in a convective environment.}
  \label{fig:fig9}
\end{figure}
 
A dense NN model architecture with 4 hidden FC layers and 10 nodes per layer was used for this case study. Trainings were performed using the Proximal Adaptive Gradient (PAG) Optimizer (ProximalAdagradOptimizer in TensorFlow[51]) based on the forward-backward splitting method[53]. A learning rate of 0.04 was used with 25,000 iterations after a grid search for hyper parameters. For regularization, LASSO regression was used with a strength of 0.001[55, 56]. Multiple models were trained using different dataset sizes ranging from 100 to 300. For the baseline training set, the ReLU activation function[52] was used. For transient heat transfer problems, it is known that the behavior of the system is exponential[3]. For example, consider the thermal response of an inert lumped mass heated via convection and a constant heating rate. The transient thermal lag follows the following exponential-decay behavior[24]:
\begin{equation}
Transient\ Lag=\Delta T_{SS} (1-e^{-\frac{T_{hold}-T_0}{\Delta T_{SS}}})
\end{equation}
In Equation 11, \(\Delta T_{SS}\) is the steady-state thermal lag[61] and \(T_{hold}-T_0\) is the difference between initial air temperature and hold temperature. This is, of course, a similar intuition to that conveyed by Newton’s law of cooling[3]. Aside from heat transfer, heat generation is also known to follow an exponential behavior.  This is also the intuition obtained from the Boltzmann distribution of molecular energies in statistical mechanics[62]. It is therefore an obvious choice to replace the ReLU activation function with an exponential based function to better capture the physics of the problem. Hence, in the second set of trainings, the hyperbolic tangent activation function (tanh[51]) was used. These activations functions are compared below: 
\begin{equation}
ReLU(z)=max(0,z) \ \ and \ \ tanh(z)=\ \frac{e^z-e^{-z}}{e^z+e^{-z}}
\end{equation}
The results from these two training sets with different activation functions are compared in Figure 10 and error values are listed under Table 2. This shows a significant improvement in the performance of the model once the activation function is switched, with RMSE reducing from 1.7 to 0.9 °C, and maximum error reducing from 6.5 to 2.7 °C.

\begin{figure}
  \centering
  \includegraphics{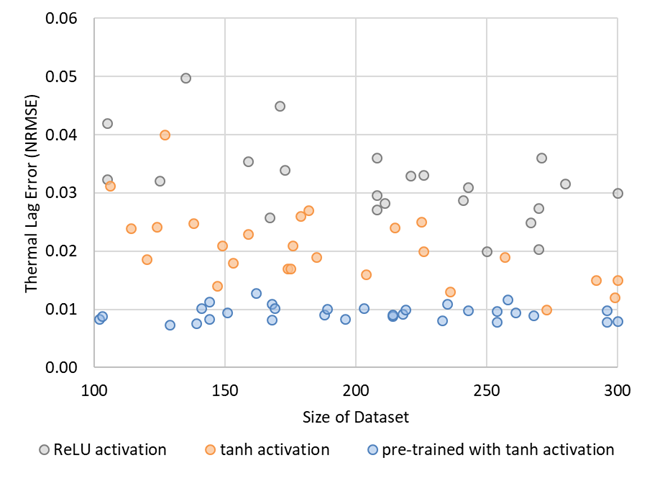}
  \caption{NRMSE of predicting thermal lag as a function of dataset size for different trainings in case study 2.}
  \label{fig:fig10}
\end{figure}

\begin{table}[ht]
\caption{Mean and max errors to predict thermal lag in case study 2 after training NN models with 300 data points and Proximal Adaptive Gradient (PAG) optimizer with a learning rate of 0.04}
\centering
\begin{tabular}{p{0.20\linewidth} | >{\centering\arraybackslash}p{0.1\linewidth} | >{\centering\arraybackslash}p{0.1\linewidth} | >{\centering\arraybackslash}p{0.1\linewidth} | >{\centering\arraybackslash}p{0.1\linewidth} | >{\centering\arraybackslash}p{0.1\linewidth} | >{\centering\arraybackslash}p{0.1\linewidth}}
\hline
\textbf{Model} & \textbf{Optimizer} & \textbf{Layers} & \textbf{Nodes/ Layer} & \textbf{NRMSE} & \textbf{RMSE (°C)} & \textbf{Max Error (°C)} \\ \hline
ReLU activation  & PAG 25K iterations & 4 & 10 & 0.030 & 1.7 & 6.5\\
tanh activation & PAG 25K iterations & 4 & 10 & 0.015 & 0.9 & 2.7\\
Pre-trained NN with tanh activation & PAG 25K iterations & 4 & 10 & 0.008 & 0.4 & 1.3\\\hline
\end{tabular}
\label{tab:table2}
\end{table}

To further improve the accuracy of the model, similar to the previous case study, one could use physics-based features in the trainings. Instead, a different approach is taken here.  In many engineering applications, approximate solutions (either closed-form or numerical) are commonly used to quickly explore the design space. Here we use two approximate closed-form solutions for the two terms in Equation 11. The first term is the steady-state thermal lag for the inert problem[61], and the second term is for the transient thermal lag of a lumped mass system[24]. Equation 11, calculated using these two approximations, is used to pre-train the NN models. Although Equation 11 captures many aspects of the underlying physics of the problem, it neglects the exothermic reaction and heat generated during cure. This approach also neglects any substantial temperature gradients through the thickness of the part. In order to effectively use this approximate solution, the following steps were taken:

\begin{itemize}
\item Step I: Firstly 8,000 data points were generated using the approximate closed-form solution.

\item Step II: An NN model was pre-trained using these 8,000 approximate data points. For training, tanh activation function, GD optimization with a learning rate of 0.001, and 100,000 iterations were used based on a grid-search of hyper parameters. 

\item Step III: The pre-trained model was then used to initialize learning for trainings with 100 to 300 exact data points. GD optimization with 25,000 iterations were used for this final step of training. 

\end{itemize}

Attention was paid to avoid data contamination between approximate and exact data points. Results obtained from these trainings are compared with results from the previous two sets of trainings (Table 2).  Pre-training of the NN model with the approximate solution improved the RMSE from 0.9 to 0.4 °C and the maximum error from 2.7 to 1.3 °C.  Again, the maximum error is brought into a range of acceptable confidence compared to typical cure cycle specifications.

This case study clearly demonstrates that using TGML, a low-fidelity model can be transformed into a high-fidelity model with a limited number of data points. It should be noted that the success of this approach depends on the capability of the low-fidelity model to capture underlying physical trends. The approximate solution used in this case study captures both the exponential decay behavior of the transient solution and also underlying trends of the steady-state solution[24, 61]. For the range of inputs used in this case study, approximate solution has a maximum error of 10 °C in predicating thermal lag in the part. This is significantly higher than the maximum error of 1.3 °C in the trained NN model. This shows that while the approximate model cannot be used for engineering applications, it can significantly guide the training of ML models.

\subsection{Case Study 3: Exotherm in a Curing Composite Part}
Here we train NN models to predict the exotherm in the problem described in the previous case study. The exotherm is defined as the maximum difference between air temperature and part temperature as schematically depicted in Figure 9. For this case study, 5,000 data points were generated using FE.  For ML, a dense NN model architecture with 4 hidden FC layers and 10 nodes per layer was used. As described before, after a grid-search, GD Optimizer with a learning rate of 0.001 and 100,000 iterations. No regularization was used in this case study. Similar to thermal lag, both ReLU and tanh activation functions were used for comparison. The correlation between error and dataset size and also the choice of activation function along with error values are listed in Table 3. With 5,000 data points, a RMSE of 0.7 °C and maximum error of 7.8 °C were obtained for models with ReLU activation. Using tanh activation, RMSE of 0.5 °C and maximum error of 4.9 °C were obtained. Although this is an improvement compared to models with ReLU activation, this is still a high error considering that in typical aerospace applications, the exotherm for this material has to be limited to 5 °C[8]. 

\begin{table}[ht]
\caption{Mean and max Errors to predict exotherm in case study 3 after training NN models with 5000 data points and GD (rate=0.001)/PAG (rate=0.06) optimizers}
\centering
\begin{tabular}{p{0.20\linewidth} | >{\centering\arraybackslash}p{0.1\linewidth} | >{\centering\arraybackslash}p{0.1\linewidth} | >{\centering\arraybackslash}p{0.1\linewidth} | >{\centering\arraybackslash}p{0.1\linewidth} | >{\centering\arraybackslash}p{0.1\linewidth} | >{\centering\arraybackslash}p{0.1\linewidth}}
\hline
\textbf{Model} & \textbf{Optimizer} & \textbf{Layers} & \textbf{Nodes/ Layer} & \textbf{NRMSE} & \textbf{RMSE (°C)} & \textbf{Max Error (°C)} \\ \hline
ReLU activation  & GD 100K iterations & 4 & 10 & 0.017 & 0.7 & 7.8\\
tanh activation & GD 100K iterations & 4 & 10 & 0.010 & 0.5 & 4.9\\
tanh activation and transformed heating rate & GD 100K iterations & 4 & 10 & 0.009 & 0.4 & 3.0\\
tanh activation and transformed heating rate & PAG 100K iterations & 4 & 10 & 0.005 & 0.2 & 1.6\\\hline
\end{tabular}
\label{tab:table3}
\end{table}

To reduce error, we introduce domain knowledge using a different approach than previous cases. Initially, based on first principles, a correlation between exotherm and heating-rate is inferred. Consider a case where all parameters in the system are fixed except for the heating rate. For a very slow heating-rate, the exotherm will be negligible, since the reaction occurs mostly during the heat-up step where part and tool have enough time to transfer the heat of reaction to the boundaries via conduction and eventually to air via convection.  As the heating rate is increased, exotherm starts to build up on the hold. As the heating rate is further increased, eventually the exotherm becomes almost constant as all of the reaction occurs during the hold. This is equivalent of a one-step isothermal heating. There are several methods that can be used to introduce such a correlation during training, for example modification of the loss function. Here a simpler approach is taken. Instead of using original values of heating rate, the following exponential transformation is performed before training:

\begin{equation}
f(T^\prime\ )=\frac{1}{1+e^{-T^\prime\ }}-0.6
\end{equation}

This transformation is a derivation of the sigmoid function as plotted in Figure 11. The constant of 0.6 in the Equation 13 is set such that the function predicts a zero value for a heating rate of about 0.5 °C/min as observed in the FE simulation. This explicitly introduces the inferred correlation by reducing the effect of heating rate using an exponential decay response. Trainings were performed using this new feature and a NN architecture as shown in Figure 11. Results are shown in Figure 12a and error values listed in Table 3. This shows a noticeable improvement in model performance as the maximum error now drops from 4.9 °C to 3.0 °C.  
 
\begin{figure}
  \centering
  \includegraphics{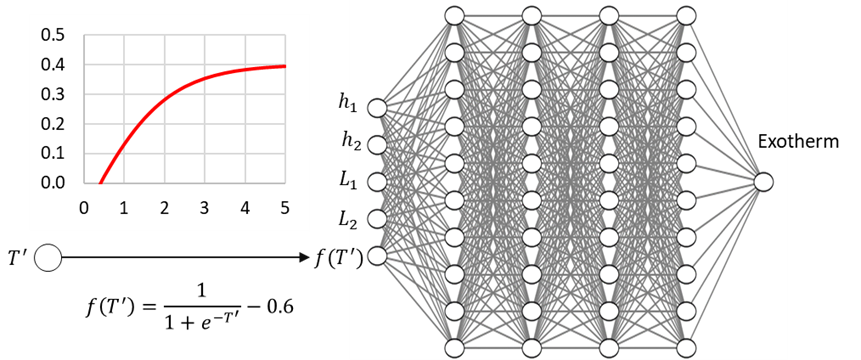}
  \caption{The NN model trained to predict exotherm in case study 3 using a physics-based transformation of heating rate based on a derivation of sigmoid function.}
  \label{fig:fig11}
\end{figure}

At this step, more correlations could be introduced to further reduce the maximum error. Instead, we focus on the capability of trained models to identify underlying trends. Consider a test case where all model parameters are kept constant except for the heating rate which is varied between 0.5 and 8 °C/min (note that trainings were performed with heating rates between 1 and 5 °C/min). A part thickness of 10 mm and a tool thickness of 12 mm were selected for evaluation. Heat transfer coefficients of 60 and 40 \(\frac{W}{m^2K}\) were set for the part and tool sides respectively. Predictions of the trained models with and without transforming heating rates are compared in Figure 12b (both with tanh activation function). Results from FE simulations are also shown in this figure. Based on Figure 12, the NN model trained with transformed heating rate can accurately predict the trend even beyond the bounds of the training zone. In contrast, the NN model which was not trained with this transformation performed very poorly outside the training zone.
The accuracy of the TGML model can be further improved by selecting a different optimization algorithm. Instead of the GD optimizer with a learning rate of 0.001, the PAG optimizer with a learning rate of 0.06 was used to train the model. Using this approach, the RMSE was reduced from 0.4 °C to 0.2 °C and maximum error value was reduced from 3 °C to 1.6 °C (Table 3).

\begin{figure}
  \centering
  \includegraphics{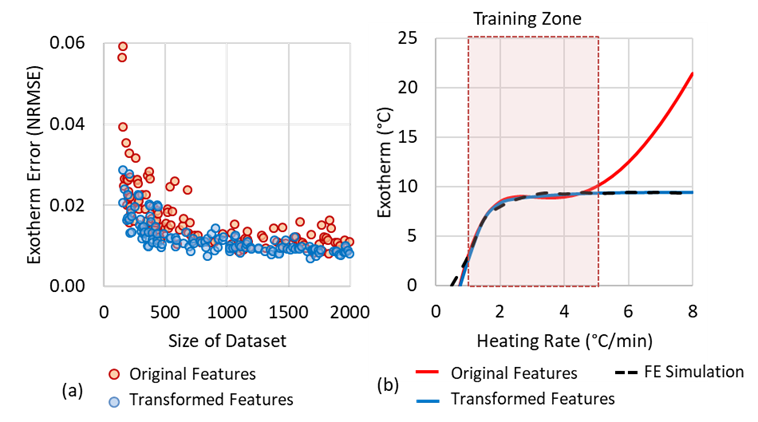}
  \caption{(a) NRMSE of predicting exotherm as a function of dataset size for trainings with original features and also using physics-based feature transformation in case study 3, and (b) predictions of the trend for exotherm using different NN models compared with finite element simulation results.}
  \label{fig:fig12}
\end{figure}

\section{Implementation and Application in Aerospace Industry}
A TGML model similar to the model in case study 3 was recently implemented in a commercial simulation tool for aerospace applications by Convergent Manufacturing Technologies (CMT). Details and some of the results are discussed in this section. During conceptual design phase of complex aerospace composite parts, it becomes critical to quickly assess the manufacturability using thermo-chemical analysis. Since design details including geometry (thickness and lay-ups) and cure cycle are not set at this step, many design iterations must be assessed. As discussed, established commercial FE tools such as COMPRO[44] in ABAQUS are widely used in industry for full 3D thermo-chemical analysis of composite parts. However, for a representative aerospace composite part such as the wing skin of a Boeing 787, a single simulation might take hours on a typical engineering workstation (e.g. i7 processor with 8 cores and no GPU) as shown in Figure 13 [25]. Consequently, design optimization might take days or weeks, which limits the exploration of the design spectrum. To increase simulation speed, CMT has developed a commercial design aid, the Composites Producibility Assessment – Thermal Analysis (CPA-TA), as a plugin for CATIA V5. CPA-TA divides a given composite part into unique zones and perform 1D FE simulations in all the zones instead of full 3D FE simulation[25]. As discussed before, for thin parts on thin tools, and away from the edges and substructures of the tool, the accuracy of a 1D simulation is acceptable in the conceptual design phase. The representative wing skin shown in Figure 13, for example, has 63 different zones which 30 of them are unique geometries[25]. Evaluating the entire wing skin with three different temperature cycles (e.g. varying heating rate) and three different tool thicknesses, requires 270 1D analyses. For comparison, analysis of a representative composite wing skin using CPA-TA and 270 1D FE simulations was conducted on a typical engineering workstation.  The entire analysis took 1620 seconds (27 minutes)[25]. This is a considerable gain in speed compared to the 3D FE simulation as shown in Figure 13. However, even with this 1D approximation, speed of simulation is such that only limited number of design cases may be explored. Based on the results of the current study, a TGML model similar to the model in case 3 was developed and implemented in CPA-TA. Aside from thermal lag and exotherm discussed in this paper, the TGML model is capable of predicting other KPIs including minimum resin viscosity during processing[25]. The same simulation was conducted on the same workstation using the commercial implementation of TGML in CPA-TA.  Similar to the previous case, the representative part was divided into 30 unique zones and 270 TGML simulations were conducted. Now, the entire analysis only took about 2 seconds. Out of this, it was determined that 1.5 seconds was used to initialize libraries in the CPA-TA code while analysis time was only 0.5 seconds. Comparison of different simulation approaches are shown in Figure 13. This shows a significant gain in simulation speed to the point of near real-time simulation capability. This opens-up tremendous opportunities for engineers to quickly explore the entire design spectrum and optimize the design.

\begin{figure}
  \centering
  \includegraphics{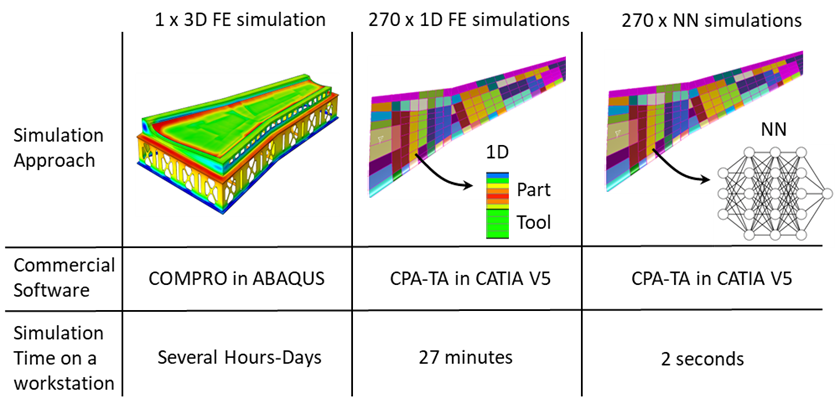}
  \caption{Comparison of simulation times using different simulation approaches for thermo-chemical analysis of a representative composite part on an Invar tool (FE simulation figures are provided by Convergent Manufacturing Technologies and used by permission in this article).}
  \label{fig:fig13}
\end{figure}

\section{Discussion}
In the context of composites processing, several techniques were introduced to incorporate theory and domain knowledge in trainings of neural networks to predict various processing related parameters. These techniques include:

\begin{itemize}
\item Physics-based feature transformation based on closed-form solution (case 1)
\item Selection of tanh activation function based on the physics of the problem (case 2)
\item Pre-training using physics-based approximate solution (case 2)
\item Physics-based feature transformation based on understanding of asymptotic behavior of solution (case 3)
\end{itemize}

These techniques are summarized in Figure 14. As the complexity of the case study is increased, more advanced techniques are used. A major advantage of TGML over ML is to ensure physical consistency in the solution as demonstrated in both case 1 and case 3. As shown in Figure 6 and Figure 12b, outside of the training zones, all ML models fail to capture the underlying correlations. TGML models on the other hand, predict a similar trend compared to the FE solution outside of the training zones. 

\begin{figure}
  \centering
  \includegraphics{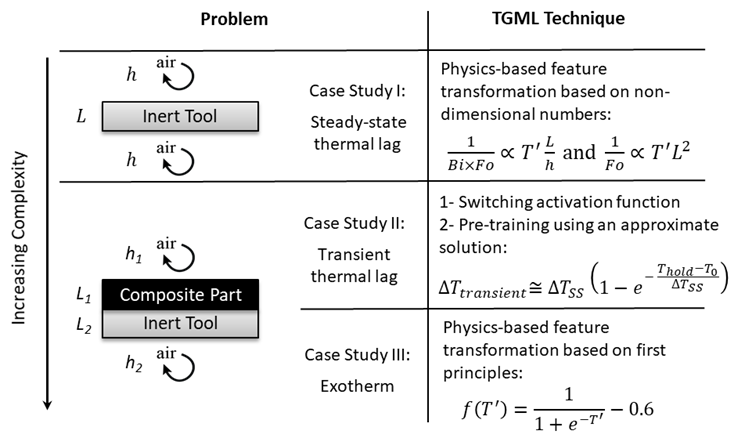}
  \caption{Summary of TGML techniques used in three case studies. As the complexity of the problem is increased, more advanced techniques are used.}
  \label{fig:fig14}
\end{figure}

The other advantage of TGML over ML is to achieve high accuracy even with small datasets and simple NN architectures. In case 1, for example, the accuracy of a small TGML model was better than the accuracy of a very complex and large ML model. The other important factor is the reduction of required data points to ensure an accurate model as demonstrated in case 2. As shown in Figure 10, even with 100 data points, using pre-training based on an approximate solution, an accurate TGML model was trained. Both small architectures and small datasets results in much faster trainings of NNs.

A major advantage of both ML and TGML modelling compared to FE modelling is speed.  For case 3, using an Intel i7-7700 CPU 3.6 GHz processor with no GPU/TPU, the mid-sized TGML 40 NN model takes 26 seconds for 100,000 simulations.  An equivalent single RAVEN 1D run takes 3-5 seconds.  This comparison indicates that a TGML model is of the order of 10,000 faster than the FE model.  Although a runtime of 3-5 seconds might be considered fast for many applications, it is not sufficient for parametric evaluations of large response surfaces nor for real-time feedback and control purposes. Effectiveness of TGML approach is shown in the recent implementation in a commercial code. The thermo-chemical analysis of a representative composite wing skin was conducted in only 2 seconds compared to 27 minutes using the traditional FE approach.

\section{Summary and Conclusions}
ML approaches are now increasingly used in scientific and physical domains thanks to a multitude of factors including IoT and exponential growth of data in industry 4.0 transition. However, due to constraints, physical data is often small and fragmented. This poses challenges to implement traditional ML approaches considering that they lack the mathematical foundation to capture complexity of physical problems especially outside their training zones or with fragmented datasets. Not surprisingly, Scientific Machine Learning (SciML) or Theory-Guided Machine Learning (TGML) to effectively incorporate domain knowledge and theory into traditional ML approaches, has been identified as one the key research areas in coming years. 

In this study, thermal processing of composites was chosen as the application for comparison of traditional ML and TGML approaches.  Three case studies were analyzed and it was shown that it is highly advantageous to integrate domain knowledge in training NN models. As the complexity of case studies was increased, theory and doing knowledge was methodically and incrementally introduced into ML trainings. Several different TGML approaches were discussed including physics-based feature transformation based on closed form solutions, selection of activation function based on the physics of the problem, pre-training using physics-based approximate solutions, and physics-based feature transformation based on understanding of asymptotic behavior of solution. It was shown that TGML becomes quite effective to reduce error in small datasets. However, even in large datasets, TGML ensures physical consistency and helps to capture underlying trends.

It was shown that although acceptable RMSE values can be obtained using traditional ML approaches, using TGML techniques, maximum error values can be reduced to the desired range quite effectively. It was also shown that by using TGML, small models with reduced number of iterations can achieve the same or even improved performance compared to large models trained using traditional ML. Beyond the training zone, none of the traditional ML approaches were capable of capturing the underlying trends. Only TGML was able to capture correct physical trends beyond the boundaries of the training zone. It should be noted that although processing of composites was used to demonstrate the capabilities of TGML, the techniques discussed here can be applied to other composites engineering applications.

\section*{Acknowledgements}
The authors would like to thank the Natural Sciences and Engineering Research Council of Canada (NSERC) and the industrial members of the Composites Research Network (The Boeing Company, Toray Americas, Convergent Manufacturing Technologies, Avcorp Industries) for their financial support. We would like to acknowledge the contributions of Dr. Anthony Floyd from Convergent Manufacturing Technologies for supporting the work, implementing the developed TGML model in CPA-TA code, and conducting simulations in section 4 to compare FE and TGML models. We would also like to acknowledge Convergent Manufacturing Technologies for providing screenshots of FE simulation results used in the current manuscript. 

\section*{References}
\begin{enumerate}
\item Agrawal A, Choudhary A (2016) Perspective: Materials informatics and big data: Realization of the “fourth paradigm” of science in materials science. APL Materials, 4(5), https://doi.org/10.1063/1.4946894
\item Zobeiry N, Humfeld KD (2019) An iterative scientific machine learning approach for discovery of theories underlying physical phenomena. arXiv, arXiv:1909
\item Incropera FP, DeWitt DP, Bergman TL, Lavine AS (2007) Fundamentals of Heat and Mass Transfer 6th Edition. Fundamentals of Heat and Mass Transfer 6th Edition, https://doi.org/10.1016/j.applthermaleng.2011.03.022
\item Surana KS, Reddy JN (2017) The finite element method for initial value problems: Mathematics and computations. The Finite Element Method for Initial Value Problems: Mathematics and Computations, https://doi.org/10.1201/b22512
\item Zobeiry N, Humfeld KD (2021) A physics-informed machine learning approach for solving heat transfer equation in advanced manufacturing and engineering applications. Engineering Applications of Artificial Intelligence, 101:104232. https://doi.org/10.1016/j.engappai.2021.104232
\item Raissi M, Perdikaris P, Karniadakis GE (2019) Physics-informed neural networks: A deep learning framework for solving forward and inverse problems involving nonlinear partial differential equations. Journal of Computational Physics, 378:686–707. https://doi.org/10.1016/j.jcp.2018.10.045
\item Sirignano J, Spiliopoulos K (2018) DGM: A deep learning algorithm for solving partial differential equations. Journal of Computational Physics, 375:1339–1364. https://doi.org/10.1016/j.jcp.2018.08.029
\item Fernlund G, Mobuchon C, Zobeiry N (2018) 2.3 Autoclave Processing. Comprehensive Composite Materials II, 2:42–62. https://doi.org/10.1016/b978-0-12-803581-8.09899-4
\item Zobeiry N, Duffner C (2018) Measuring the negative pressure during processing of advanced composites. Composite Structures, 203:11–17. https://doi.org/10.1016/j.compstruct.2018.06.123
\item Zobeiry N, Forghani A, Li C, Gordnian K, Thorpe R, Vaziri R, Fernlund G, Poursartip A (2016) Multiscale characterization and representation of composite materials during processing. Philosophical Transactions of the Royal Society A: Mathematical, Physical and Engineering Sciences, 374(2071):20150278. https://doi.org/10.1098/rsta.2015.0278
\item Zobeiry N, Poursartip A (2015) The origins of residual stress and its evaluation in composite materials. Structural Integrity and Durability of Advanced Composites: Innovative Modelling Methods and Intelligent Design, :43–72. https://doi.org/10.1016/B978-0-08-100137-0.00003-1
\item Li C, Zobeiry N, Keil K, Chatterjee S, Poursartip A (2014) Advances in the characterization of residual stress in composite structures. International SAMPE Technical Conference, 
\item Farnand K, Zobeiry N, Poursartip A, Fernlund G (2017) Micro-level mechanisms of fiber waviness and wrinkling during hot drape forming of unidirectional prepreg composites. Composites Part A: Applied Science and Manufacturing, 103:168–177. https://doi.org/10.1016/j.compositesa.2017.10.008
\item Farhang L, Mohseni M, Zobeiry N, Fernlund G (2020) Experimental study of void evolution in partially impregnated prepregs. Journal of Composite Materials, 54(11):1511–1523. https://doi.org/10.1177/0021998319883934
\item Carl H. Z, Peter B (2018) Comprehensive Composite Materials II. Comprehensive Composite Materials II, IIhttps://www.medicalpdfbooks.org/comprehensive-composite-materials-ii.pdf
\item Fernlund G, Poursartip A, Nelson K, Wilenski M, Swanstrom F (1999) Process modeling for dimensional control-sensitivity analysis of a composite spar process. International SAMPE Symposium and Exhibition (Proceedings), 
\item Fabris J, Lussier D, Zobeiry N, Mobuchon C, Poursartip A (2014) Development of standardized approaches to thermal management in composites manufacturing. International SAMPE Technical Conference, 
\item Zobeiry N, Vaziri R, Poursartip A (2006) Differential implementation of the viscoelastic response of a curing thermoset matrix for composites processing. Journal of Engineering Materials and Technology, Transactions of the ASME, 128(1):90–95. https://doi.org/10.1115/1.2148421
\item Fabris J, Mobuchon C, Zobeiry N, Lussier D, Fernlund G, Poursartip A (2015) Introducing thermal history producibility assessment at conceptual design. International SAMPE Technical Conference, 2015-Janua
\item Zobeiry N, Vaziri R, Poursartip A (2015) Characterization of strain-softening behavior and failure mechanisms of composites under tension and compression. Composites Part A: Applied Science and Manufacturing, 68:29–41. https://doi.org/10.1016/j.compositesa.2014.09.009
\item Mohseni M, Zobeiry N, Fernlund G (2019) Experimental and numerical study of coupled gas and resin transport and its effect on porosity. Journal of Reinforced Plastics and Composites, 38(23–24):1055–1066. https://doi.org/10.1177/0731684419865783
\item Johnston A, Hubert P, Fernlund G, Vaziri R, Poursartip A (1996) Process modelling of composite structures employing a virtual autoclave concept. Science and Engineering of Composite Materials, 
\item Fernlund G, Osooly A, Poursartip A, Vaziri R, Courdji R, Nelson K, George P, Hendrickson L, Griffith J (2003) Finite element based prediction of process-induced deformation of autoclaved composite structures using 2D process analysis and 3D structural analysis. Composite Structures, https://doi.org/10.1016/S0263-8223(03)00117-X
\item Zobeiry N, Park J, Poursartip A (2019) An infrared thermography-based method for the evaluation of the thermal response of tooling for composites manufacturing. Journal of Composite Materials, 53(10):1277–1290. https://doi.org/10.1177/0021998318798444
\item Zobeiry N, VanEe D, Anthony F, Poursartip A (2019) Theory-guided machine learning composites processing modelling for manufacturability assessment in preliminary design. In NAFEMS 17th World Congress, 
\item Fabris J, Mobuchon C, Zobeiry N, Poursartip A (2016) Understanding the consequences of tooling design choices on thermal history in composites processing. International SAMPE Technical Conference, 2016-Janua
\item Park J, Zobeiry N, Poursartip A (2017) Tooling materials and their effect on surface thermal gradients. International SAMPE Technical Conference, :2554–2568. 
\item Fernlund G, Rahman N, Courdji R, Bresslauer M, Poursartip A, Willden K, Nelson K (2002) Experimental and numerical study of the effect of cure cycle, tool surface, geometry, and lay-up on the dimensional fidelity of autoclave-processed composite parts. Composites - Part A: Applied Science and Manufacturing, https://doi.org/10.1016/S1359-835X(01)00123-3
\item Erol S, Schuhmacher A, Sihn W (2016) Strategic guidance towards industry 4.0 - a three-stage process model. International Conference on Competitive Manufacturing (COMA), :495–500. 
\item Singh AP, Medida S, Duraisamy K (2017) Machine-learning-augmented predictive modeling of turbulent separated flows over airfoils. AIAA Journal, 55(7):2215–2227. https://doi.org/10.2514/1.J055595
\item Wang J-XX, Wu J-LL, Xiao H (2017) Physics-informed machine learning approach for reconstructing Reynolds stress modeling discrepancies based on DNS data. Physical Review Fluids, 2(3):034603. https://doi.org/10.1103/PhysRevFluids.2.034603
\item Belianinov A, He Q, Kravchenko M, Jesse S, Borisevich A, Kalinin S V. (2015) Identification of phases, symmetries and defects through local crystallography. Nature Communications, 6https://doi.org/10.1038/ncomms8801
\item Karpatne A, Atluri G, Faghmous JH, Steinbach M, Banerjee A, Ganguly A, Shekhar S, Samatova N, Kumar V (2017) Theory-guided data science: A new paradigm for scientific discovery from data. IEEE Transactions on Knowledge and Data Engineering, 29(10):2318–2331. https://doi.org/10.1109/TKDE.2017.2720168
\item Wagner N, Rondinelli JM (2016) Theory-guided machine learning in materials science. Frontiers in Materials, 3https://doi.org/10.3389/fmats.2016.00028
\item Zobeiry N, Reiner J, Vaziri R (2020) Theory-guided machine learning for damage characterization of composites. Composite Structures, 246:112407. https://doi.org/10.1016/j.compstruct.2020.112407
\item Liang L, Liu M, Martin C, Sun W (2018) A deep learning approach to estimate stress distribution: a fast and accurate surrogate of finite-element analysis. Journal of the Royal Society Interface, https://doi.org/10.1098/rsif.2017.0844
\item Lee CW, Rice BP (1996) Modeling of epoxy cure reaction rate by neural network. International SAMPE Technical Conference, 
\item Lee CW, Gibson T, Tienda KA, Storage TM (2010) Reaction rate and viscosity model development for Cytec’s Cycom® 5320 family of resins. International SAMPE Technical Conference, 
\item Nelson KM, Poursartip A, Fernlund G (2000) Cure kinetics and the dimensional control of composite structure. International SAMPE Symposium and Exhibition (Proceedings), 
\item Hubert P, Johnston A, Poursartip A, Nelson K (2001) Cure kinetics and viscosity models for Hexcel 8552 epoxy resin. International SAMPE Symposium and Exhibition (Proceedings), 
\item Fernlund G, Nelson K, Poursartip A (2000) Modeling of process induced deformations of composite shell structures. International SAMPE Symposium and Exhibition (Proceedings), 
\item Hubert P, Fernlund G, Poursartip A (2012) Autoclave processing for composites. Manufacturing Techniques for Polymer Matrix Composites (PMCs), https://doi.org/10.1016/B978-0-85709-067-6.50013-4
\item Ee D Van, Poursartip A (2009) HexPly 8552 material properties database for use with COMPRO CCA and RAVEN, created for NCAMP. 
\item Www.convergent.ca/products/compro-simulation-software (2014) COMPRO simulation software. www.convergent.ca/products/compro-simulation-software
\item Www.esi-group.com/software-solutions/virtual-manufacturing/composites/pam-composites (2018) PAM-Composites. www.esi-group.com/software-solutions/virtual-manufacturing/composites/pam-composites
\item Www.convergent.ca/products/raven-simulation-software (2013) RAVEN simulation software. 
\item Slesinger NA (2010) Thermal modeling validation techniques for thermoset polymer matrix composites. University of British Columbia, (July)https://doi.org/10.14288/1.0071063
\item HEXCEL (2016) HexPly 8552 Epoxy matrix (180°C/365°F curing matrix) Product Data. 
\item Hexcel (2015) HexTow ® AS4. Material Datasheet, 
\item Fabris J (2018) A framework for formalizing science based composites manufacturing practice. https://doi.org/10.14288/1.0372787
\item Abadi M, Agarwal A, Barham P, Brevdo E, Chen Z, Citro C, Corrado GS, Davis A, Dean J, Devin M, Ghemawat S, Goodfellow I, Harp A, Irving G, Isard M, Jia Y, Jozefowicz R, Kaiser L, Kudlur M, Levenberg J, Mane D, Monga R, Moore S, Murray D, Olah C, Schuster M, Shlens J, Steiner B, Sutskever I, Talwar K, Tucker P, Vanhoucke V, Vasudevan V, Viegas F, Vinyals O, Warden P, Wattenberg M, Wicke M, Yu Y, Zheng X (2016) TensorFlow: Large-Scale Machine Learning on Heterogeneous Distributed Systems. http://arxiv.org/abs/1603.04467
\item Nair V, Hinton GE (2010) Rectified linear units improve Restricted Boltzmann machines. ICML 2010 - Proceedings, 27th International Conference on Machine Learning, 
\item Duchi J, Singer Y (2009) Efficient learning using forward-backward splitting. Advances in Neural Information Processing Systems 22 - Proceedings of the 2009 Conference, 
\item Kingma DP, Ba JL (2015) Adam: A method for stochastic optimization. 3rd International Conference on Learning Representations, ICLR 2015 - Conference Track Proceedings, 
\item Tibshirani R (1996) Regression Shrinkage and Selection Via the Lasso. Journal of the Royal Statistical Society: Series B (Methodological), https://doi.org/10.1111/j.2517-6161.1996.tb02080.x
\item Friedman J, Hastie T, Tibshirani R (2008) Sparse inverse covariance estimation with the graphical lasso. Biostatistics, https://doi.org/10.1093/biostatistics/kxm045
\item Glorot X, Bengio Y (2010) Understanding the difficulty of training deep feedforward neural networks. Journal of Machine Learning Research, 
\item Rasekh A, Vaziri R, Poursartip A (2004) Simple techniques for thermal analysis of the processing of composite structures. 36th International SAMPE Tech Conference, 
\item Jolliffe IT (2002) Principal Component Analysis, Second Edition. Encyclopedia of Statistics in Behavioral Science, https://doi.org/10.2307/1270093
\item Rai N, Pitchumani R (1997) Rapid cure simulation using artificial neural networks. Composites Part A: Applied Science and Manufacturing, 28(9–10):847–859. https://doi.org/10.1016/S1359-835X(97)00046-8
\item Rasekh A (2007) Efficient Methods for Non-Linear Thermochemical Analysis of Composite Structures Undergoing Autoclave Processing. The University of British Columbia, https://doi.org/10.14288/1.0063235
\item Chandler D (1987) Introduction to modern statistical mechanics. Introduction to Modern Statistical Mechanics. 
\end{enumerate}
\end{document}